\pdfoutput=1
\documentclass{article}
\usepackage{spconf,amsmath,graphicx}
\usepackage{tabularx}
\usepackage{multirow}
\usepackage{booktabs}
\usepackage{amsmath}

\title{DRSM: EFFICIENT NEURAL 4D DECOMPOSITION FOR DYNAMIC RECONSTRUCTION IN STATIONARY MONOCULAR CAMERAS}
\name{Weixing Xie\textsuperscript{1,2,3,\dag}, Xiao Dong\textsuperscript{4,\dag }, Yong Yang\textsuperscript{1}, Qiqin Lin\textsuperscript{1}, Jingze Chen\textsuperscript{1}, Junfeng Yao\textsuperscript{1,2,3,*}, Xiaohu Guo\textsuperscript{5}\thanks{*Corresponding author. \dag Equal contribution. The research was supported by Natural Science Foundation of China (No.62072388) and public technology service platform project of Xiamen city (No.3502Z20231043).}}
    \address{\textsuperscript{1}Center for Digital Media Computing, School of Film, School of Informatics, Xiamen University\\\textsuperscript{2}National Institute for Data Science in Health and Medicine, Xiamen University\\\textsuperscript{3}Key Laboratory of Digital Protection and Intelligent Processing of Intangible Cultural Heritage\\of Fujian and Taiwan, Ministry of Culture and Tourism
    \\\textsuperscript{4}Department of Computer Science, BNU-HKBU United International College\\\textsuperscript{5}Department of Computer Science, The University of Texas at Dallas}
\begin{document}
%
\maketitle
\begin{abstract}
With the popularity of monocular videos generated by video sharing and live broadcasting applications, reconstructing and editing dynamic scenes in stationary monocular cameras has become a special but anticipated technology. In contrast to scene reconstructions that exploit multi-view observations, the problem of modeling a dynamic scene from a single view is significantly more under-constrained and ill-posed. Inspired by recent progress in neural rendering, we present a novel framework to tackle 4D decomposition problem for dynamic scenes in monocular cameras. 
Our framework utilizes decomposed static and dynamic feature planes to represent 4D scenes and emphasizes the learning of dynamic regions through dense ray casting. Inadequate 3D clues from a single-view and occlusion are also particular challenges in scene reconstruction. To overcome these difficulties, we propose deep supervised optimization and ray casting strategies. With experiments on various videos, our method generates higher-fidelity results than existing methods for single-view dynamic scene representation.

\end{abstract}

\section{introduction}
\label{sec:intro}

In recent years, the popularity of short videos and live broadcasts has led to the generation of a lot of video data, most of which are dynamic content from a single perspective of a fixed camera. We try to efficiently reconstruct and realistically render dynamic scenes in single view videos. Dynamic scenes in the video may be disturbed or obscured by other objects, such as hands (Fig. \ref{Fig2}) and wires (Fig. \ref{Fig3}). 
Our goal is to accurately recover the entire static and dynamic scene of interest to the viewer while removing occluding objects.
\begin{figure}
\centering
 \includegraphics[width=1.0\linewidth]{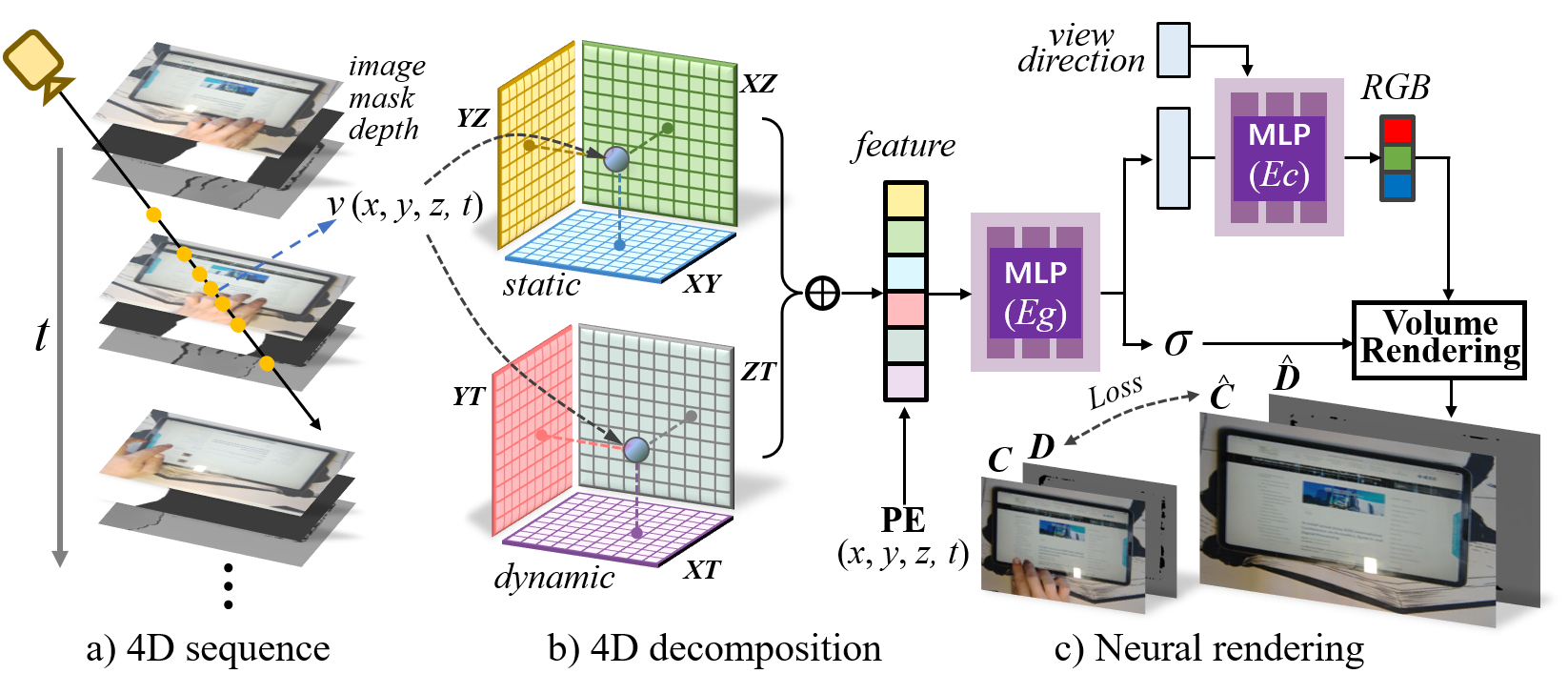}\\[-1em]
  \caption{Framework of the proposed DRSM.}
  \vspace{-15pt}
  \label{Fig1}
\end{figure}
Neural Radiation Fields (NeRF) \cite{mildenhall2020nerf} tackles novel view synthesis of static scene by learning implicit representations of objects from multiple captured views.  To model dynamic scenes, many works \cite{gao2022monocular} propose the ray deformation paradigms that parameterizes a deformed scene as a NeRF in canonical space with a time-dependent deformation for dynamic reconstruction \cite{pumarola2021d, park2021nerfies, gao2021dynamic, li2021neural, tretschk2021non}. Other works learn the 4D scene representation by decoupling static and dynamic scenes with different NeRFs \cite{wu2022d, zhang2023detachable}. For example, D$^{2}$NeRF \cite{wu2022d} achieves dynamic and static decoupling, which can remove all dynamic objects in the scene.
But this does not solve the problem where we want to reconstruct dynamic and static scenes simultaneously.

Typically, dynamic NeRFs rely on video flow captured by multi-view cameras \cite{peng2021neural, li2022neural} or one free-viewpoint camera \cite{pumarola2021d, park2021nerfies, gao2021dynamic, li2021neural, tretschk2021non} to get full view perception of dynamic scenes. Different with them, we aim to solve the problem of modeling dynamic scene in single view, which is ill-posed and challenging due to limited geometric perception.
Many works exploit auxiliary information to help understand the structure of the scene, such as SMPL \cite{SMPL:2015} prior to help constrain human motion space~\cite{weng2022humannerf, zhao2022high} or depth prior to help recover geometry of objects~\cite{xian2021space,cai2022neural}. Among them, NDR \cite{xian2021space} solves the geometric reconstruction of moving objects and can be modified as a background reconstruction technique to solve our problem. 

Moreover, the optimization for dynamic NeRFs is computationally intensive since it requires multiple MLP evaluations. To avoid huge memory footprint of previous methods~\cite{yu2021plenoctrees}, we decouple spatial and temporal features via planar factorization~\cite{chan2022efficient} to model 4D field for single-view videos.

Overall, our technical contributions are as follows: 1) We propose an efficient 4D decomposition framework (DRSM) with planar factorization for fast \textbf{D}ynamic \textbf{R}econstruction in \textbf{S}tationary \textbf{M}onocular Cameras; 2) we address the inherent motion-appearance ambiguity for single-view using depth prior; 3) we propose an efficient importance sampling strategy (ISDM) based on dynamic and mask regions to improve the reconstruction quality for time-variant and occluded regions; 4) we demonstrate a convincing rendering quality and smooth point clouds on multiple short-form videos.

\section{Method}
\label{sec:format}

The architecture of our network DRSM is shown in Fig.~\ref{Fig1}. We take a video $V = \{\mathbf{I}_i, \mathbf{D}_i, \mathbf{M}_i : i \in [1, T]\}$ from a single viewpoint as input, where $\mathbf{I}_i$ is the $i$-th frame image, $\mathbf{D}_i$ is the corresponding depth image and $\mathbf{M}_i$ is the mask of occluded objects to be removed. The object mask can be obtained by combining the Segment Anything Method (SAM)  \cite{kirillov2023segany} with the OSTrack tracking model \cite{ye2022ostrack}. The video duration is normalized to [0, 1]. Thus, time of the $i$-th frame is $i/T$. 

Our network starts by randomly picking a frame for training. We employ the ISDM sampling strategy to identify high-priority region and build casting rays. For sampling points along casting ray, we use bilinear interpolation to query their features on spatial and temporal tri-planes and construct the fused features, which are then passed to MLP decoders to predict color and density. We apply volume rendering to generate color and depth for each casting ray, and design rendering losses for supervision. After training, the network learns 4D representation and can reconstruct video, point cloud and synthesize novel views.

\subsection{Preliminaries}
\label{2.1}
NeRF \cite{mildenhall2020nerf} learns a regression function $\textit{F}$ that takes the encoded coordinates of a 3D point $\textbf{x} = (\textit{x, y, z})$ observed from a view direction $\textbf{d} = (\theta, \phi)$ as input, and outputs the corresponding radiance \textbf{c} and volume density $\sigma$: $\textit{F}_{\text{NeRF}}: (\mathbf{x}, \mathbf{d}) \rightarrow (\mathbf{c}, \sigma)$. 
The estimated color $\mathit{\hat{C}}(\mathbf{r})$ and depth  $\mathit{\hat{D}}(\mathbf{r})$ of a pixel can be rendered by integrating the radiance by tracking a ray $\mathbf{r}(s) = \mathbf{o} + s\mathbf{d}$, cast from the camera toward the center of the pixel:
\begin{equation}\label{eq1}
\begin{split}
\hat{{C}}(\mathbf{r}) = \int_{s_{n}}^{s_{f}} T(s)\sigma(\mathbf{r}(s))\mathbf{c}(\mathbf{r}(s),\mathbf{d}),
\end{split}
\end{equation}
\begin{equation}\label{eq2}
\begin{split}
\hat{D}(\mathbf{r}) = \int_{s_{n}}^{s_{f}} T(s)\sigma(\mathbf{r}(s))s\,ds,
\end{split}
\end{equation}
\begin{equation}\label{eq3}
\begin{split}
T(s) = \exp\left(- \int_{s_{n}}^{s} \sigma(\mathbf{r}(p)) \ dp\right).
\end{split}
\end{equation}
$T(s)$ is the accumulated transmittance along the ray $\textbf{r}$ up to $s$.

\subsection{4D decomposition for dynamic scenes}
\label{2.2}

A dynamic scene could be naively represented as a 4D volume $\mathbf{V}$. Inspired by \cite{cao2023hexplane}, we  decompose \textbf{V} into a static volume and a dynamic volume by planar factorization:
\begin{equation}\label{eq4}
\begin{split}
\mathbf{V} = \{\mathbf{V}_s\{P_{XY}, P_{XZ}, P_{YZ}\}, \mathbf{V}_d\{P_{XT}, P_{YT}, P_{ZT}\}\}.
\end{split}
\end{equation}

\noindent
Here static volume $\mathbf{V}_s$ is projected to a tri-plane representing only spaces of $xy$, $xz$, and $yz$. The dynamic volume $\mathbf{V}_d$ is projected to a tri-plane representing spaces and time, $xt$, $yt$, and $zt$. Each plane has dimension $N \times N \times W$, where $N$ is the resolution and $W$ is the number of feature channels. This approach allows us to represent a 4D volume efficiently using six planes (Fig. \ref{Fig1}b). For a 4D point $v=(x, y, z, t)$, we can query its features $\mathbf{f}(v)$ by projecting it onto these planes and use bilinear interpolation $\psi$ to obtain the corresponding values:
\begin{equation}\label{eq5}
\begin{split}
\mathbf{f}(v) = \psi (P_{XY}, x, y) \odot \psi (P_{XZ}, x, z) \odot \psi (P_{YZ}, y, z)\\\odot \psi (P_{XT}, x, t) \odot \psi (P_{YT}, y, t) \odot \psi (P_{ZT}, z, t),
\end{split}
\end{equation}
where $\psi (P_{XY}, x, y)$ means given regularly spaced feature plane $P_{XY}$ and the $x,y$ coordinates, using bilinear interpolation to calculate the plane feature of $v$. The $\odot$ represents Hadamard product to get fused features.

We use two small MLPs (Fig. \ref{Fig1}c) to decode the fused features $\mathbf{f}(v)$ like Instant-NGP \cite{muller2022instant}. The features and positional encoding are concatenated and fed into the geometry MLP $E_g$ to obtain density $\sigma$ and high dimensional features $\mathbf{f}^{\prime}(v)$:
\begin{equation}\label{eq6}
\begin{split}
\sigma(v), \mathbf{f}^{\prime}(v) = E_g(\mathbf{f}(v), \gamma(v)).
\end{split}
\end{equation}
Here, $\gamma(\cdot)$ is an encoding function \cite{mildenhall2020nerf}. Then, we concatenate the feature $\mathbf{f}^{\prime}(v)$ with the positional encoding of view direction $(\theta, \phi)$ and feed it into the color MLP $E_c$ to obtain the radiance:
\begin{equation}\label{eq7}
\begin{split}
\mathbf{c}(r, g, b) = E_c(\mathbf{f}^{\prime}(v), \gamma(\theta, \phi)).
\end{split}
\end{equation}

\begin{figure*}
  \centering
  \includegraphics[width=0.96\linewidth]{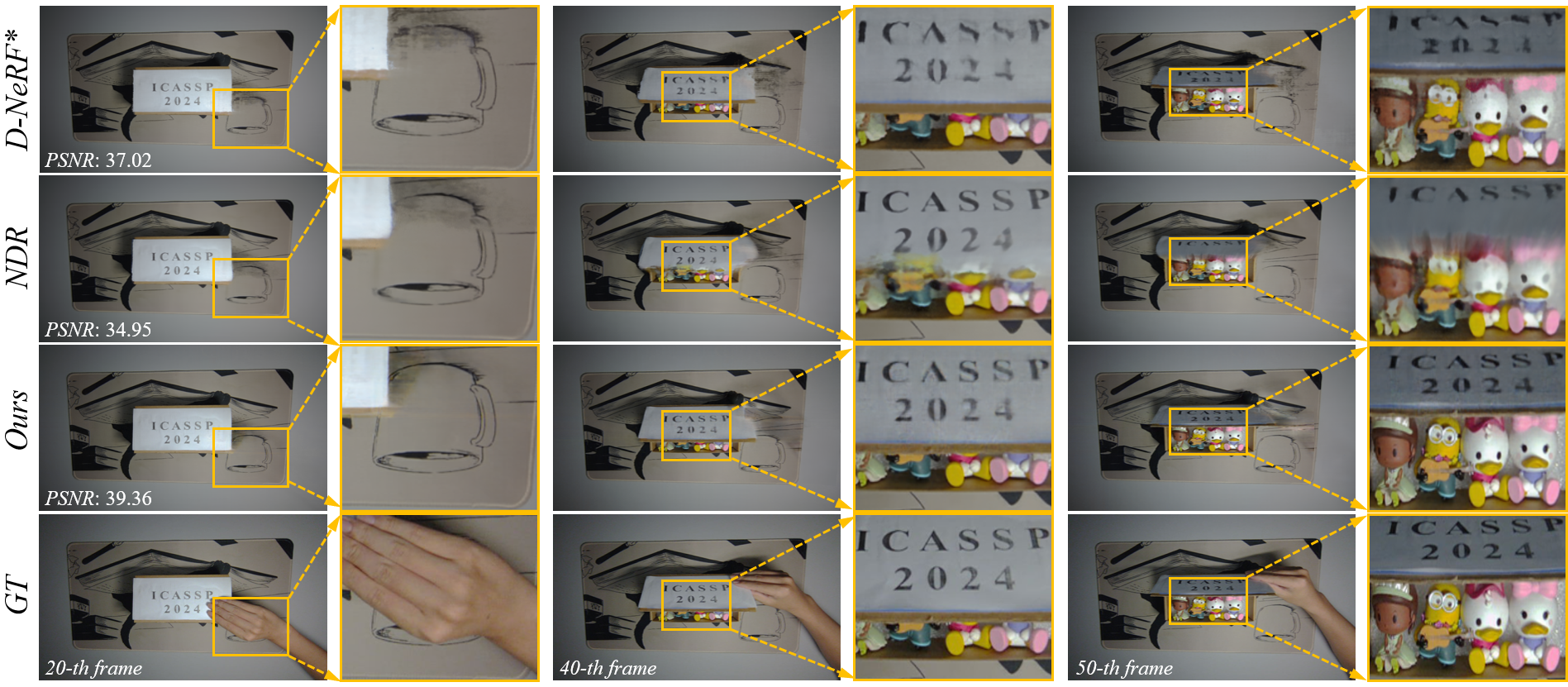}\\[-1em]
  \caption{Comparison of DRSM and other methods on dynamic reconstruction results. We remove the hand in video and show \textit{PSNR} metric of each method.}
  \label{Fig2}
\end{figure*}

\subsection{ISDM sampling strategy}
\label{2.3}
Previous scene representation methods \cite{gao2022monocular} usually randomly sample a batch of pixels/rays on the whole input image for training. In our work, we focus on learning the representation for dynamic scene of interest while removing occluding objects. Uniform sampling is no longer suitable for our method because dynamic areas and occluded areas require higher sampling weights. 

We propose the importance sampling strategy based on dynamic and mask regions. For the occlusion mask $\mathbf{M}_i$ of frame $i$ (0 for occluded pixels), we ignore those pixels in occluded region in the ray selection. We create an importance map $\tilde{\mathbf{P}}_{i}$ to guide the pixel sampling, assigning higher probability for those regions with higher occlusion frequencies across all frames. The sampling importance map is calculated according to element-wise division:
\begin{equation}\label{eq8}
\begin{split}
\tilde{\mathbf{P}}_{i} = {\mathbf{M}_{i}T}/({\sum_{k=1}^{T} \mathbf{M}_{k} + \boldsymbol{\varepsilon}}).
\end{split}
\end{equation}
In addition to occlusion areas, we should also prioritize sampling dynamic areas. In uniform sampling, a large proportion of selected pixels may fall into the static background, which contributes less to the dynamic reconstruction. To identify the dynamic region, we calculate temporal difference of pixels on frames $i$ and $j$~\cite{li2022neural}:
\begin{equation}\label{eq9}
\begin{split}
\mathbf{P}_{i} = \tilde{\mathbf{P}}_{i} \odot \min(\frac{1}{3} \left\| \mathbf{I}_i - \mathbf{I}_j \right\|_1, \alpha), j \in (i - \tau, i + \tau),
\end{split}
\end{equation}
where $\alpha$ is a lower-bound parameter controlling the sampling weights of the dynamic region and $\tau$ is set to 25 in the experiment. ISDM sampling adjusts the sampling probability of time-varying and occlusion areas, which helps improve reconstruction quality and speed up training.

\subsection{Optimization}
\label{2.4}
We supervise scene reconstruction in terms of reconstructed image $\hat{C}$, depth $\hat{D}$, and regularization loss to optimize the parameters of feature planes and MLPs. For each batch of training data, there are $R$ rays sampled by ISDM strategy on one frame. We first minimize the difference between the ground truth color and the predicted color, as shown in Eq.~\eqref{eq1}. To assist in scene representation for single-view input, we further optimize the geometry using depth supervision. The color loss and depth loss are shown in the following equations:
\begin{equation}\label{eq10}
\begin{split}
\mathcal{L}_{\text{color}} = \frac{1}{|\mathcal{R}|} \sum_{\mathbf{r} \in \mathcal{R}} \left\| C(\mathbf{r}) - \hat{C}(\mathbf{r}) \right\|_2^2,
\end{split}
\end{equation}
\begin{equation}\label{eq11}
\begin{split}
\mathcal{L}_{\text{depth}} = \frac{1}{|\mathcal{R}|} \sum_{\mathbf{r} \in \mathcal{R}} \left\| D(\mathbf{r}) - \hat{D}(\mathbf{r}) \right\|_2^2.
\end{split}
\end{equation}
Dynamic scene reconstruction in stationary monocular camera is a severely ill-posed problem. To achieve robust reconstruction, we apply strong regularizers. We adopt 2D total variation (TV) loss $\mathcal{L}_{\text{TV-2D}}$ for space planes in \cite{muller2022instant} and 1D TV loss $\mathcal{L}_{\text{TV-1D}}$ on the space axis for space-time planes and a similar smooth loss $\mathcal{L}_{\text{smooth}}$ on the time axis. The total optimization objective is:
\begin{equation}\label{eq12}
\begin{split}
\mathcal{L} = \mathcal{L}_{\text{color}} + \lambda_{\text{1}}\mathcal{L}_{\text{depth}} + \lambda_{\text{2}}\mathcal{L}_{\text{2D}} + \lambda_{\text{3}}\mathcal{L}_{\text{1D}} + \lambda_{\text{4}}\mathcal{L}_{\text{smooth}}.
\end{split}
\end{equation}

\begin{figure*}
\centering
 \includegraphics[width=0.95\linewidth]{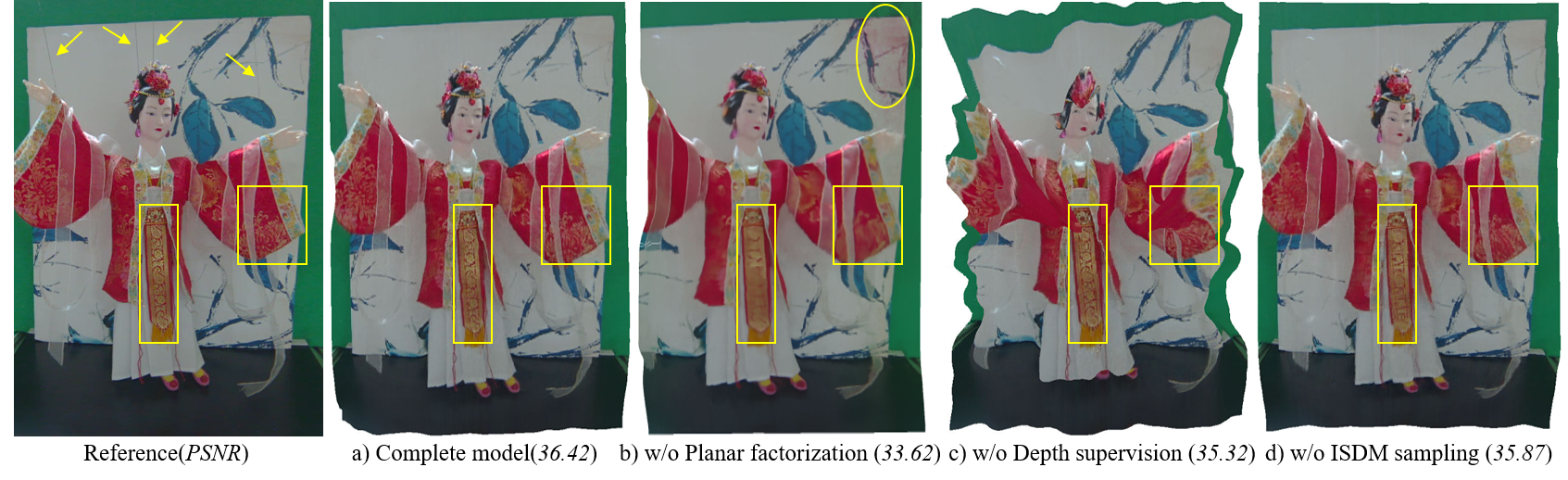}\\[-1em]
  \caption{Ablation study on a marionette dancing video. We remove manipulating wires and show the reconstructed point clouds.}
  \label{Fig3}
\end{figure*}

\begin{table*}
\footnotesize
\setlength\tabcolsep{2pt}
\centering
\caption{Quantitative comparisons on our collected dataset. We report PSNR$\uparrow$, SSIM$\uparrow$, LPIPS$\downarrow$ and training time (minutes).}
\begin{tabular}{lllllllllllllllllllllll} 
\toprule
\multirow{2}{*} {Model} & \multicolumn{3}{c}{\textquotedblleft Box\textquotedblright} & \multicolumn{3}{c}{\textquotedblleft Marionette\textquotedblright} & \multicolumn{3}{c}{\textquotedblleft Web page\textquotedblright} & \multicolumn{3}{c}{\textquotedblleft Xiangqi\textquotedblright} & \multicolumn{3}{c}{\textquotedblleft Calligraphy\textquotedblright} & \multicolumn{3}{c}{\textquotedblleft Toy\textquotedblright} & \multirow{2}{*} {Time} \\
\cmidrule(lr){2-4} \cmidrule(lr){5-7} \cmidrule(lr){8-10} \cmidrule(lr){11-13} \cmidrule(lr){14-16} \cmidrule(lr){17-19} 
   &PSNR & SSIM & LPIPS &PSNR & SSIM & LPIPS &PSNR & SSIM & LPIPS &PSNR & SSIM & LPIPS &PSNR & SSIM & LPIPS &PSNR & SSIM & LPIPS\\
   \midrule
  D-NeRF\textsuperscript{*} & 37.02 & 0.945 & 0.084 & 33.62 & 0.885 & 0.058 & 35.80 & 0.947 & 0.068 & 35.21 & 0.959 & 0.049 & 36.27 & 0.911 & 0.095 & 37.14 & 0.944 & 0.088 & 477min \\
  NDR   & 34.95 & 0.942 & 0.094 & 33.74 & 0.909 & 0.053 & 35.37 & 0.944 & 0.086 & 35.24 & 0.955 & 0.062 & 35.02 & 0.890 & 0.139 & 35.03 & 0.940 & 0.097 & 666min \\
  Ours-5K & 38.07 & 0.945 & 0.071 & 34.12 & 0.918 & 0.044 & 37.79 & 0.954 & 0.047 & 35.20 & 0.930 & 0.048 & 37.20 & 0.928 & 0.073 & 37.05 & 0.942 & 0.089 & \textbf{15min} \\
  Ours-10K & \textbf{39.36} & \textbf{0.953} & \textbf{0.070} & \textbf{36.42} & \textbf{0.945} & \textbf{0.023} & \textbf{39.38} & \textbf{0.964} & \textbf{0.043} & \textbf{37.27} & \textbf{0.964} & \textbf{0.037} & \textbf{38.51} & \textbf{0.943} & \textbf{0.058} & \textbf{38.42} & \textbf{0.950} & \textbf{0.079} & 35min \\
  \bottomrule
  \end{tabular}\\[1em]
  
  \label{table1}
\end{table*}

\section{experiments}
\label{sec:format}

\textbf{Experimental settings.} We normalize the scene into device coordinates (NDC) to handle monocular videos and then sample casting rays within the NDC space. We use a model with four symmetric spatial resolutions 64, 128, 256 and 512. The feature length $W$ at each scale is set to 32. We set the frequencies of positional encoding $\gamma(\cdot)$ for sampling points and view direction to 4. In each training iteration, a batch contains \(\mathcal{R} = 2048\) sampling rays. The loss weights in Eq.\eqref{eq12} are empirically set as \(\lambda_1 = 1.0\), \(\lambda_2 = 0.0002\), \(\lambda_3 = 0.0001\), \(\lambda_4 = 0.001\). Adam \cite{adam} optimizer is adopted for training, and the initial learning rate is set to \(0.01\). We train all scenes with \(5k\) and \(10k\) iterations on a single RTX 3090 GPU, which take around \(15\) and \(35\) minutes, respectively. We build a video dataset, including life videos related to box, marionette, web page, xiangqi, calligraphy and toy. We use the ``Record3D'' app on iPhone and RGBD camera to record videos. Each video lasts for \(5\sim7\) seconds and we sample $10$ frames per second for training.


\textbf{Comparison experiments.} We compare our method with other dynamic scene reconstruction methods for monocular videos, such as D-NeRF \cite{pumarola2021d} and NDR \cite{cai2022neural}. D-NeRF builds a deformable neural radiance field based on a canonical 3D representation and time-guided motion fields. However, the model performance of D-NeRF depending on a canonical frame suffers when objects exhibit long-distance translations \cite{ramasinghe2023bali}. NDR focuses on modeling dynamic foreground objects based on bijective motion map and implicit  representations of MLPs. Using an MLP with a specific bandwidth to learn both spatial and temporal variations simultaneously results in suboptimal reconstruction of complex scenes.

As shown in Fig.~\ref{Fig2}, D-NeRF\textsuperscript{*} and NDR are modified version of original models with depth supervision for fair comparison with our method. D-NeRF\textsuperscript{*} failed to capture the deformation of long-distance moving objects, i.e., the characters on the box. The predicted color of the doll predicted by NDR is affected by the movement of the box. This is because NDR's bijective map focuses on learning the geometric changes of moving objects not the high-frequency details of static part. Our network is specifically designed for the reconstruction of combined static and dynamic scenes, resulting in better video appearance reconstructions. 

In Table~\ref{table1}, we show the indicators such as PSNR, SSIM and LPIPS of 6 videos to quantitatively evaluation the reconstruction. Our model outperforms the existing methods on multiple aspects and requires shorter training time.

\textbf{Ablation study.} We present ablation experiments on network modules and the reconstructed point clouds in Fig.~\ref{Fig3}. The marionette dancing video contains some manipulating wires to be removed. Without planar factorization, our network failed to reconstruct high quality texture details in static region (color prediction error in yellow ellipse) as well as dynamic region (sleeves and decorations in yellow box). Furthermore, we observe severe distortions in the reconstructed point cloud when depth supervision is disabled, indicating that the network is unable to learn the correct geometry from single-view input without prior. Without ISDM sampling, predicted high-frequency textures also become blurry. We provide the corresponding PSNR indicator to further demonstrate the effectiveness of proposed modules. Our complete model produces high-fidelity reconstructions.


\section{CONCLUSION}
\label{sec:format}

This paper presents a novel neural 4D decomposition for dynamic reconstruction from single-view videos. Without observation from multi-viewpoints, the problem of modeling dynamic scenes is typically quite challenging. We apply planar decomposition to static and dynamic scenes respectively to improve the model's modeling ability of 4D scenes. To address the ambiguous geometry, we utilize depth prior to constrain the motion space. The adaptive sampling strategies aid the reconstruction on moving objects and occluding regions. The ablation study demonstrates the effectiveness of proposed network. We conduct rich experiments to show the superiority of our network than existing methods on the special task of single-view dynamic scene construction.

\newpage


\bibliographystyle{IEEEbib}
\bibliography{refs}

\end{document}